%% file: main.tex
\documentclass[preprint,12pt]{elsarticle}
\usepackage[T1]{fontenc}
\usepackage[utf8]{inputenc}
\usepackage{amsmath,amssymb}
\usepackage{graphicx}
\usepackage{booktabs}
\usepackage{adjustbox}
\usepackage{algorithm}
\usepackage{algpseudocode}
\usepackage{xcolor}
\usepackage{pifont}
\usepackage{textcomp}
\usepackage{fontawesome5}
\usepackage{tikz}
\usetikzlibrary{positioning,arrows.meta,shadows,calc}
\usepackage[hidelinks]{hyperref}
\usepackage{xurl}
\usepackage{float}  %
\usepackage{xspace}
\newcommand{\Method}{SCOUT\xspace} %
\input{generated/numbers}
\journal{Computer Methods and Programs in Biomedicine}
\newcommand{\MainTabMain}{\ref{tab:main}}\newcommand{\MainTabLoso}{\ref{tab:loso}}

\begin{document}
\begin{frontmatter}
\title{\Method: spending a foundation model where tissue detection fails}

\author[dieti]{Cristian Tommasino\corref{cor}}
\ead{cristian.tommasino@unina.it}
\author[dieti]{Antonio M. Rinaldi}
\ead{antoniomaria.rinaldi@unina.it}
\cortext[cor]{Corresponding author.}
\affiliation[dieti]{organization={Department of Electrical Engineering and Information Technology (DIETI), University of Naples Federico II},
            city={Naples},
            country={Italy}}

\begin{abstract}
\noindent\textbf{Background and objective:} Tissue detection is the first step of almost every whole-slide image analysis, and its errors pass silently downstream. Released detectors are judged by mean overlap with manual masks, which can hide missed pale tissue or fragments. We asked how often their masks need correction, and whether running a pathology foundation model only where a cheap detector is likely wrong makes detection more reliable on unseen sources.

\noindent\textbf{Methods:} On the public Tissues with Issues benchmark we compared, under an operational correction criterion, five released detectors (threshold and post-processing tuned on development data), a U-Net trained on the same masks, and \Method. \Method sends the thumbnail windows that a small audit network ranks riskiest to a frozen foundation model, fuses by consensus, and flags slides with a calibrated score. We held out each of five sources in turn and tested ten external slides.

\noindent\textbf{Results:} Even after tuning, released detectors failed on \FailReleasedTMin--\FailHistoqcDevtuned\,\% of the case groups, mostly by missing immunohistochemistry tissue. \Method failed on \FailTrace\,\%, like the U-Net (\FailUnet\,\%). On held-out sources \Method failed on \LosoFailTrace\,\%, the U-Net on \LosoFailUnet\,\% and the re-tuned HEST on \LosoFailHestT\,\%; both comparisons hinged on one source. On external slides, HEST failed \BandiFailHestT{} of \NumBandi, \Method \BandiFailTrace. The review flag caught \FlagSensTen\,\% of the failed case groups.

\noindent\textbf{Conclusions:} Released tissue detectors need correction far more often than their overlap suggests. Training on target-convention masks removes most failures on familiar sources. With one configuration, \Method had the lowest failure rate, alone or tied, on familiar, held-out and external slides; it flags slides for review and is released as a verified, pip-installable package.
\end{abstract}

\end{frontmatter}

\input{sections_cmpb/introduction}
\input{sections/related}
\input{sections/method}
\input{sections_cmpb/experiments}
\input{sections_cmpb/software}
\input{sections_cmpb/results}
\input{sections_cmpb/discussion}

\section*{CRediT authorship contribution statement}
\textbf{Cristian Tommasino:} Conceptualization, Methodology, Software, Validation, Formal analysis, Investigation, Data curation, Writing -- original draft, Visualization. \textbf{Antonio M. Rinaldi:} Conceptualization, Supervision, Writing -- review \& editing.

\section*{Declaration of competing interest}
The authors declare that they have no known competing financial interests or personal relationships that could have appeared to influence the work reported in this paper.

\section*{Funding}
This research did not receive any specific grant from funding agencies in the public, commercial, or not-for-profit sectors.

\section*{Data and code availability}
All datasets are public: Tissues with Issues, the B\'andi et al.\ test slides and CAMELYON17, used under their licences; no ethics approval was required. The \texttt{scout-tissue} package, the weights of \Method-open and an archived release are available at \url{https://github.com/CosmoIknosLab/scout-tissue-detection}, \url{https://pypi.org/project/scout-tissue/} and \href{https://doi.org/10.5281/zenodo.23664072}{doi:10.5281/zenodo.23664072}.

\section*{Declaration of generative AI and AI-assisted technologies in the manuscript preparation process}
During the preparation of this work the authors used Claude (Anthropic) as an assistant to write and debug analysis code, run and check experiments, and draft and edit the text. After using this tool, the authors reviewed and edited the content as needed and take full responsibility for the content of the published article.

\bibliographystyle{elsarticle-num}
\bibliography{references}
\clearpage
\section*{Supplementary Material}
\setcounter{section}{0}\renewcommand{\thesection}{S\arabic{section}}
\setcounter{table}{0}\renewcommand{\thetable}{S\arabic{table}}
\setcounter{figure}{0}\renewcommand{\thefigure}{S\arabic{figure}}

\input{sections/supplement_body}

\end{document}

%% file: generated/numbers.tex
\newcommand{\NumTestSlides}{153}  %
\newcommand{\NumTestGroups}{113}  %
\newcommand{\FailUnet}{19.5}  %
\newcommand{\FailHestDevtunedT}{29.2}  %
\newcommand{\DiceHestDevtunedT}{0.979}  %
\newcommand{\DiffTraceHestDevtunedT}{-10.6}  %
\newcommand{\DiffTraceHestDevtunedTLo}{-20.4}  %
\newcommand{\DiffTraceHestDevtunedTHi}{-0.9}  %
\newcommand{\DiffTraceAtlaspatchDevtunedTAbs}{18.6}  %
\newcommand{\FailHestDevtuned}{47.8}  %
\newcommand{\FailHistoqcDevtuned}{94.7}  %
\newcommand{\AdjGtGlass}{8.2}  %
\newcommand{\MaskMpp}{8}  %
\newcommand{\FmHeadThr}{0.18}  %
\newcommand{\FmFpTarget}{5}  %
\newcommand{\FmEpochs}{30}  %
\newcommand{\FmBatch}{64}  %
\newcommand{\FmLr}{1e-3}  %
\newcommand{\FmWd}{1e-4}  %
\newcommand{\AuditCropPx}{256}  %
\newcommand{\AuditBatch}{16}  %
\newcommand{\AuditLr}{1e-3}  %
\newcommand{\AuditWd}{1e-4}  %
\newcommand{\WinPx}{56}  %
\newcommand{\WinUm}{448}  %
\newcommand{\FmMpp}{2}  %
\newcommand{\FmInputPx}{224}  %
\newcommand{\NearMm}{0.5}  %
\newcommand{\AuditParamsM}{1.4}  %
\newcommand{\AuditEpochs}{40}  %
\newcommand{\BudgetPct}{50}  %
\newcommand{\FusionCut}{0.5}  %
\newcommand{\MinCompMm}{0.01}  %
\newcommand{\HoleMm}{0.25}  %
\newcommand{\BandUm}{16}  %
\newcommand{\FailEps}{2}  %
\newcommand{\FragMinMm}{0.1}  %
\newcommand{\FragRecall}{0.5}  %
\newcommand{\FailFp}{10}  %
\newcommand{\FailPrec}{0.9}  %
\newcommand{\CertDelta}{0.05}  %
\newcommand{\CertAlpha}{10}  %
\newcommand{\CertAlphaFive}{5}  %
\newcommand{\NBoot}{10{,}000}  %
\newcommand{\NumDevSlides}{217}  %
\newcommand{\NumCalSlides}{185}  %
\newcommand{\NumDevGroups}{161}  %
\newcommand{\NumCalGroups}{128}  %
\newcommand{\BaseOd}{0.06}  %
\newcommand{\BaseCloseUm}{32}  %
\newcommand{\BaseHoleMm}{1}  %
\newcommand{\NumTuneCombos}{180}  %
\newcommand{\NumBandi}{10}  %
\newcommand{\NumBandiStrict}{7}  %
\newcommand{\RtTraceGpu}{33.3}  %
\newcommand{\RtHestGpu}{65.0}  %
\newcommand{\RtGrandqcGpu}{1.0}  %
\newcommand{\RtTraceGpuAllMedian}{35}  %
\newcommand{\RtTraceGpuAllPninety}{76}  %
\newcommand{\RtTraceFmShare}{88}  %
\newcommand{\FailTrace}{18.6}  %
\newcommand{\CertTraceFlagTen}{31.4}  %
\newcommand{\CertTraceSilentTen}{11}  %
\newcommand{\CertTraceFlagFive}{71.2}  %
\newcommand{\CertTraceSilentFive}{3}  %
\newcommand{\BandiFailTrace}{1}  %
\newcommand{\BandiFailTraceStrict}{0}  %
\newcommand{\SixAucTrace}{0.964}  %
\newcommand{\SixLostTrace}{0.03}  %
\newcommand{\SixLostGrandqc}{0.00}  %
\newcommand{\SixLostClam}{3.11}  %
\newcommand{\SixAucDiffAtlaspatchLo}{-0.005}  %
\newcommand{\SixAucDiffGrandqcLo}{-0.007}  %
\newcommand{\SixNSlides}{500}  %
\newcommand{\SixNAnnotated}{50}  %
\newcommand{\SixMargin}{0.01}  %
\newcommand{\SixPatchUm}{112}  %
\newcommand{\SevenFailHopt}{19.5}  %
\newcommand{\RbRandomN}{50}  %
\newcommand{\RbRandFlagSilent}{16}  %
\newcommand{\UnetFailMean}{18.9}  %
\newcommand{\UnetFailSd}{1.0}  %
\newcommand{\NumSeedsTrace}{5}  %
\newcommand{\SeedFailMean}{20.4}  %
\newcommand{\SeedFailSd}{1.4}  %
\newcommand{\LosoNGroups}{107}  %
\newcommand{\NumLosoSources}{5}  %
\newcommand{\LosoFailTrace}{28.0}  %
\newcommand{\LosoFailUnet}{38.3}  %
\newcommand{\SingNGroups}{91}  %
\newcommand{\SingFailTrace}{18.7}  %
\newcommand{\SingFailUnet}{20.9}  %
\newcommand{\SingDiffUnet}{-2.2}  %
\newcommand{\SingDiffUnetLo}{-8.8}  %
\newcommand{\SingDiffUnetHi}{3.3}  %
\newcommand{\SingFailAtlaspatch}{38.5}  %
\newcommand{\SingDiffAtlaspatch}{-19.8}  %
\newcommand{\SingDiffAtlaspatchLo}{-29.7}  %
\newcommand{\SingDiffAtlaspatchHi}{-9.9}  %
\newcommand{\SingFailHest}{28.6}  %
\newcommand{\SingDiffHest}{-9.9}  %
\newcommand{\SingDiffHestLo}{-20.9}  %
\newcommand{\SingDiffHestHi}{1.1}  %
\newcommand{\FlagSensTen}{61.9}  %
\newcommand{\FlagSensFive}{95.2}  %
\newcommand{\FlagGroupsFive}{69.0}  %
\newcommand{\LosoMacroTrace}{25.0}  %
\newcommand{\CandOutsideMaxPct}{0.02}  %
\newcommand{\OracleSelTest}{3.5}  %
\newcommand{\BandiStrictUnet}{1}  %
\newcommand{\LosoFailCnnTier}{38.3}  %
\newcommand{\LosoSeedDiffUnet}{-10.3}  %
\newcommand{\LosoSeedDiffUnetLo}{-16.2}  %
\newcommand{\LosoSeedDiffUnetHi}{-4.4}  %
\newcommand{\AugSixFail}{29.5}  %
\newcommand{\AugSixLoso}{36.4}  %
\newcommand{\AugSixBFail}{23.9}  %
\newcommand{\AugSixBLoso}{36.8}  %
\newcommand{\AugSixBLosoDiff}{-8.7}  %
\newcommand{\AugSixBLosoDiffLo}{-14.0}  %
\newcommand{\AugSixBLosoDiffHi}{-3.7}  %
\newcommand{\StainIhcFailTrace}{22.0}  %
\newcommand{\StainIhcFailReleasedMin}{28.0}  %
\newcommand{\LosoNoAcrobatNGroups}{77}  %
\newcommand{\LosoNoAcrobatFailTrace}{13.0}  %
\newcommand{\LosoNoAcrobatFailUnet}{19.0}  %
\newcommand{\LosoNoAcrobatDiffUnet}{-6.1}  %
\newcommand{\LosoNoAcrobatDiffUnetLo}{-12.6}  %
\newcommand{\LosoNoAcrobatDiffUnetHi}{0.0}  %
\newcommand{\LosoNoAcrobatPUnet}{0.06}  %
\newcommand{\LosoNoAcrobatDiffCnn}{-6.5}  %
\newcommand{\LosoNoAcrobatDiffCnnLo}{-11.7}  %
\newcommand{\LosoNoAcrobatDiffCnnHi}{-1.3}  %
\newcommand{\LosoNoOtherDiffUnetMin}{-16.7}  %
\newcommand{\LosoNoOtherDiffUnetMax}{-8.8}  %
\newcommand{\LosoNoOtherDiffUnetHiMax}{-3.1}  %
\newcommand{\LosoNoAcrobatFailHestT}{31.2}  %
\newcommand{\LosoNoAcrobatDiffHestT}{-18.2}  %
\newcommand{\LosoNoAcrobatDiffHestTLo}{-28.6}  %
\newcommand{\LosoNoAcrobatDiffHestTHi}{-7.8}  %
\newcommand{\LosoNoOtherDiffHestTMin}{-3.1}  %
\newcommand{\LosoNoOtherDiffHestTMax}{6.2}  %
\newcommand{\CNineNT}{9}  %
\newcommand{\CNineTMin}{0.05}  %
\newcommand{\CNineTMax}{0.8}  %
\newcommand{\CNineTHest}{0.3}  %
\newcommand{\PTraceHestDevtunedT}{0.038}  %
\newcommand{\CalFailHestDevtunedT}{28.1}  %
\newcommand{\BandiFailHestT}{9}  %
\newcommand{\BandiStrictHestT}{6}  %
\newcommand{\LosoFailHestT}{29.9}  %
\newcommand{\LosoDiffHestT}{-1.9}  %
\newcommand{\LosoDiffHestTLo}{-13.1}  %
\newcommand{\LosoDiffHestTHi}{9.3}  %
\newcommand{\LosoPHestT}{0.795}  %
\newcommand{\LosoMacroHestT}{32.7}  %
\newcommand{\LosoAcrobatFailHestT}{26.7}  %
\newcommand{\CNineTAtlaspatch}{0.2}  %
\newcommand{\BandiStrictAtlaspatchT}{3}  %
\newcommand{\LosoFailAtlaspatchT}{38.3}  %
\newcommand{\LosoDiffAtlaspatchT}{-10.3}  %
\newcommand{\LosoDiffAtlaspatchTLo}{-17.8}  %
\newcommand{\LosoDiffAtlaspatchTHi}{-2.8}  %
\newcommand{\BandiStrictHelgesenT}{4}  %
\newcommand{\BandiStrictGrandqcT}{3}  %
\newcommand{\CalFailTrace}{9.4}  %
\newcommand{\LosoAcrobatFailTrace}{66.7}  %
\newcommand{\FailReleasedTMin}{29.2}  %
\newcommand{\PkgParitySlides}{18}  %
\newcommand{\PkgParityTwi}{15}  %
\newcommand{\PkgParityBandi}{3}  %
\newcommand{\PkgParityMaxDiffPx}{0}  %

%% file: sections_cmpb/introduction.tex
\section{Introduction}\label{sec:intro}

\begin{figure*}[t]\centering
\input{figures/fig_overview}
\caption{\Method on a Ki67 slide. The audit head selects high-risk windows for frozen-FM inference; consensus fusion refines the mask, and a calibrated score determines review. Images are contrast-enhanced for display.}
\label{fig:pipeline}
\end{figure*}
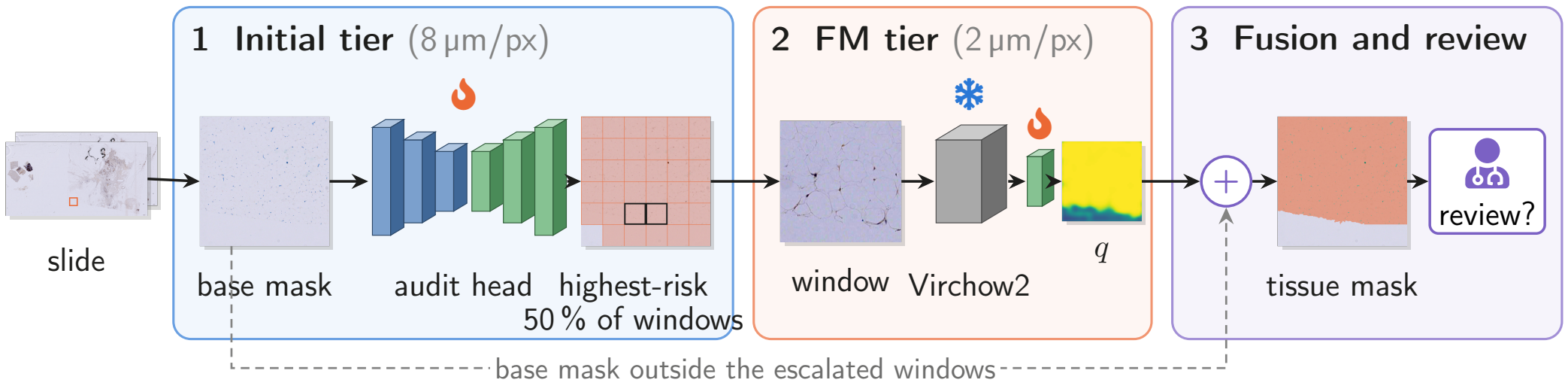

Tissue detection is the first step of most computational pathology pipelines. Its errors propagate undetected: omitted biopsy fragments are excluded from downstream analysis, whereas glass, debris, and pen marks introduce irrelevant inputs~\cite{boman2026}. High mean overlap with manual masks on H\&E~\cite{bandi2019,weng2024grandqc} has made thresholding methods and pretrained detectors the defaults of widely used toolkits~\cite{lu2021clam,zhang2025trident}.

Mean overlap, however, can hide the omission of pale immunohistochemistry (IHC) tissue, small fragments and faintly stained regions. A user does not experience a mean Dice; they experience a mask that has to be corrected or not. We therefore measure detectors by the frequency of masks that \emph{require correction}, using an explicit operational criterion at the case-group level, on a public benchmark with manual masks from many sources and stains~\cite{polejowska2026twi}.

This paper asks three practical questions. How often do released detectors need correction under this criterion, after their threshold and post-processing have been adapted to the benchmark? How much of the remaining error is removed simply by training a segmentation network on masks drawn in the same convention? And can a pathology foundation model (FM), whose representations may transfer across stains and scanners, be used at a bounded cost to make detection more reliable on sources that were not seen in training? Running an FM on every window of a slide is expensive, and an unconstrained FM tends to call unstained glass and debris tissue.

We present \Method (Selective Correction Of Uncertain Tissue), an open-source cascade that uses a learned audit of a cheap thumbnail mask to send only the riskiest windows to a frozen FM (Fig.~\ref{fig:pipeline}). Its components are individually known; the contribution is their combination into a validated, efficient tool and the evidence on when it helps. Specifically:
\begin{itemize}
\item a benchmark of five released detectors and two threshold-based pipelines under an operational correction criterion, showing that they fail on \FailReleasedTMin--\FailHistoqcDevtuned\,\% of the test case groups even after their threshold and post-processing are re-tuned, mostly by missing IHC tissue;
\item a controlled comparison with a U-Net trained on the same masks and with the re-tuned detectors, showing that adaptation to the annotation convention removes most failures on sources represented in training, and that on held-out sources the cascade degrades less than the U-Net and matches the best re-tuned detector, with both comparisons depending largely on one source;
\item \Method, a risk-guided cascade that processes only the highest-risk half of the candidate windows, constrains new tissue through consensus fusion, and flags slides for review with a threshold calibrated by Learn-then-Test; with a single configuration it has the lowest failure rate, alone or tied, on represented sources, held-out sources and an external set;
\item \texttt{scout-tissue}, a pip-installable Python package with trained weights, a command-line tool, a Python interface and a browser demo, verified to reproduce the masks of the reference implementation.
\end{itemize}

%% file: figures/fig_overview.tex
\input{figures/overview/picks}
\definecolor{cA}{HTML}{2A78D6}\definecolor{cB}{HTML}{EB6834}\definecolor{cC}{HTML}{7B61C4}
\definecolor{ink}{HTML}{1D1D1B}\definecolor{encC}{HTML}{4F81BD}\definecolor{decC}{HTML}{5DAE6E}\definecolor{fmC}{HTML}{8C8C8C}
\newcommand{\cuboid}[6]{\begin{scope}[shift={({#1},{#2})}]
  \fill[#6!75!white, draw=#6!55!black, line width=0.3pt] (0,0) rectangle (#3,#4);
  \fill[#6!45!white, draw=#6!55!black, line width=0.3pt] (0,#4) -- ++(#5,0.6*#5) -- ++(#3,0) -- ++(-#5,-0.6*#5) -- cycle;
  \fill[#6!80!black, draw=#6!55!black, line width=0.3pt] (#3,0) -- ++(#5,0.6*#5) -- ++(0,#4) -- ++(-#5,-0.6*#5) -- cycle;
\end{scope}}
\resizebox{\textwidth}{!}{%
\begin{tikzpicture}[x=1cm, y=1cm, font=\sffamily\small, >={Stealth[length=2mm,width=1.8mm]},
  tile/.style={inner sep=0pt, draw=ink!60, line width=0.4pt, drop shadow={opacity=0.3, shadow xshift=0.5mm, shadow yshift=-0.5mm}},
  lbl/.style={font=\sffamily\small, text=ink, align=center, anchor=north},
  inf/.style={->, line width=0.9pt, draw=ink},
  ttl/.style={font=\sffamily\bfseries, text=ink, anchor=north west, inner sep=0pt}]
\fill[cA!6, rounded corners=6pt] (2.25,0) rectangle (9.0,4.0);   \draw[cA!70, line width=0.8pt, rounded corners=6pt] (2.25,0) rectangle (9.0,4.0);
\fill[cB!6, rounded corners=6pt] (9.25,0) rectangle (14.05,4.0); \draw[cB!70, line width=0.8pt, rounded corners=6pt] (9.25,0) rectangle (14.05,4.0);
\fill[cC!7, rounded corners=6pt] (14.3,0) rectangle (19.0,4.0);  \draw[cC!70, line width=0.8pt, rounded corners=6pt] (14.3,0) rectangle (19.0,4.0);
\node[ttl] at (2.45,3.8) {1 \;Initial tier \textcolor{ink!55}{\textmd{(8\,\textmu m/px)}}};
\node[ttl] at (9.45,3.8) {2 \;FM tier \textcolor{ink!55}{\textmd{(2\,\textmu m/px)}}};
\node[ttl] at (14.5,3.8) {3 \;Fusion and review};
\node[tile] at (1.2,2.05) {\includegraphics[width=1.7cm]{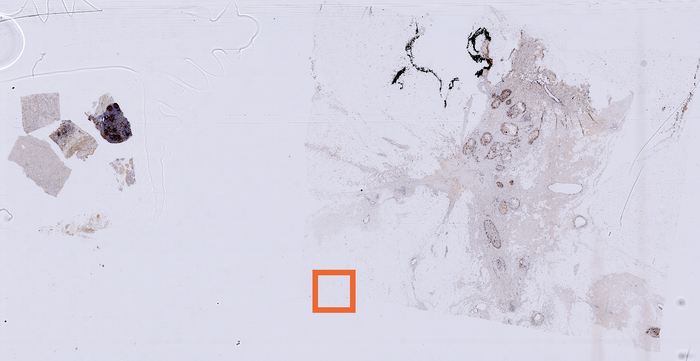}};
\node[tile] (wsi) at (1.08,1.93) {\includegraphics[width=1.7cm]{figures/overview/thumb.png}};
\node[lbl] at (1.08,1.25) {slide};
\node[tile] (base) at (3.35,1.9) {\includegraphics[width=1.55cm]{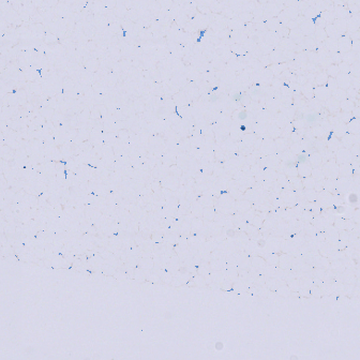}};
\node[lbl] at (3.35,0.95) {base mask};
\foreach \i/\h in {0/1.3, 1/1.0, 2/0.72} {\cuboid{4.65+\i*0.38}{1.25+(1.3-\h)/2}{0.22}{\h}{0.16}{encC}}
\foreach \i/\h in {0/0.72, 1/1.0, 2/1.3} {\cuboid{5.85+\i*0.38}{1.25+(1.3-\h)/2}{0.22}{\h}{0.16}{decC}}
\node[font=\small, text=cB] at (5.75,2.95) {\faFire};
\node[lbl] at (5.75,0.95) {audit head};
\node[tile] (risk) at (7.95,1.9) {\includegraphics[width=1.55cm]{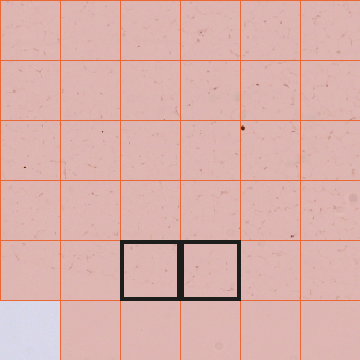}};
\node[lbl] at (7.8,0.95) {highest-risk\\[-2pt]50\,\% of windows};
\draw[inf] (wsi.east) -- (base.west);
\draw[inf] (base.east) -- (4.6,1.9);
\draw[inf] (7.0,1.9) -- (risk.west);
\node[tile] (crop) at (10.3,1.9) {\includegraphics[width=1.45cm]{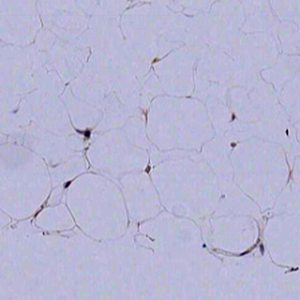}};
\node[lbl] at (10.3,1.0) {window};
\cuboid{11.45}{1.4}{0.55}{1.0}{0.3}{fmC}
\node[font=\small, text=cA] at (11.85,2.95) {\faSnowflake};
\node[lbl] at (11.85,0.95) {Virchow2};
\cuboid{12.55}{1.6}{0.18}{0.6}{0.14}{decC}
\node[font=\small, text=cB] at (12.7,2.6) {\faFire};
\node[tile] (q) at (13.45,1.9) {\includegraphics[width=0.95cm]{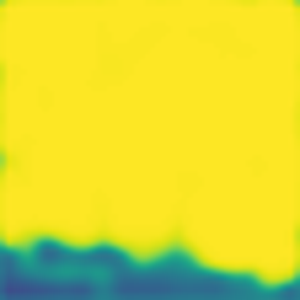}};
\node[lbl] at (13.45,1.3) {$q$};
\draw[inf] (risk.east) -- (crop.west);
\draw[inf] (crop.east) -- (11.43,1.9);
\draw[inf] (12.32,1.9) -- (12.53,1.9);
\draw[inf] (12.86,1.9) -- (q.west);
\node[circle, draw=cC, line width=0.9pt, fill=white, minimum size=6mm, inner sep=0pt, font=\sffamily\bfseries\small, text=cC] (fuse) at (14.95,1.9) {+};
\node[tile] (fin) at (16.35,1.9) {\includegraphics[width=1.55cm]{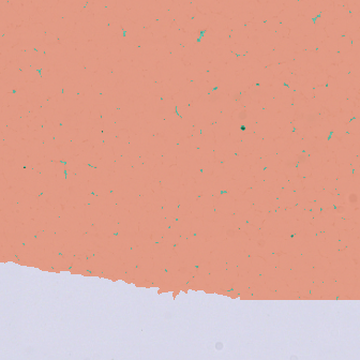}};
\node[lbl] at (16.35,0.95) {tissue mask};
\node[draw=cC, line width=0.9pt, rounded corners=3pt, fill=white, align=center, font=\sffamily\small, inner sep=3pt] (flag) at (18.1,1.9) {\textcolor{cC}{\Large\faUserMd}\\[1pt]review?};
\draw[inf] (q.east) -- (fuse.west);
\draw[inf] (fuse.east) -- (fin.west);
\draw[inf] (fin.east) -- (flag.west);
\draw[->, line width=0.7pt, draw=ink!55, densely dashed] (2.95,1.12) -- (2.95,-0.35) -- (14.95,-0.35) -- (fuse.south);
\node[font=\sffamily\footnotesize, text=ink!65, fill=white, inner sep=1.5pt] at (9.15,-0.35) {base mask outside the escalated windows};
\end{tikzpicture}}

%% file: sections/related.tex
\section{Related work}\label{sec:related}

\emph{Classical detectors.} Most pipelines threshold grayscale or saturation with Otsu's method~\cite{otsu1979} and refine the resulting masks using morphological operations~\cite{lu2021clam}; entropy-based and two-pass approaches have also been proposed~\cite{song2023entropy, ceachi2025doublepass}. Tuned on the target data, such methods achieve performance comparable to learned detectors on H\&E~\cite{bandi2017, bandi2019}, but their parameters rarely transfer across stains and scanners.

\emph{Learned detectors and quality control.} HistoQC~\cite{janowczyk2019histoqc} combines classical modules into a usable-tissue mask. GrandQC~\cite{weng2024grandqc}, the HEST tissue model~\cite{jaume2024hest} and AtlasPatch~\cite{alagha2026atlaspatch} are networks trained on large cohorts; HEST and AtlasPatch are default detectors in TRIDENT~\cite{zhang2025trident}. Helgesen et al.~\cite{helgesen2026} train nnU-Net on nearly 30,000 images of varied tissues and scanners and release the model. Independent evaluation of these models against multi-stain reference annotations has been limited; the Tissues-with-Issues benchmark~\cite{polejowska2026twi} enables such evaluation, and Boman et al.~\cite{boman2026} show that infrequent detection failures can affect downstream diagnoses.

\emph{Deferral and cascades.} Learning to defer and selective prediction route uncertain inputs to an expert or abstain~\cite{mozannar2020l2d,geifman2017selective,dvijotham2023codoc}; DeferredSeg extends deferral to pixels of medical segmentations with jointly trained expert branches~\cite{tian2027deferredseg}. Cascades and spatially adaptive inference allocate additional computation to regions where a computationally efficient model is uncertain~\cite{wang2022committees,kirillov2020pointrend}. Segmentation quality control estimates the reliability of segmentation outputs~\cite{robinson2018rtqc,devries2018leveraging, jungo2019assessing,zenk2025failure}, but generally does not incorporate a corrective inference stage. \Method combines these ideas: a learned audit of the initial mask determines where a frozen foundation model~\cite{zimmermann2024virchow2,saillard2024hoptimus0,chen2024uni} is consulted, under a fixed per-slide budget.

\emph{Risk control.} Conformal methods bound per-image errors or quality ranges of segmentations~\cite{angelopoulos2024crc,wundram2024conformal} and flag unreliable predictions in pathology and radiology~\cite{olsson2022conformal,angelopoulos2024triage}. Learn-then-Test~\cite{angelopoulos2025ltt} calibrates decision thresholds to control a specified risk with high probability; we use it to bound the rate of silent failures of a slide-level review flag, a bound that is marginal over case groups drawn i.i.d. from the benchmark's sources. Jain et al.~\cite{jain2026vlm} query a vision-language model for every slide patch and distil its predictions, without an explicit inference budget or risk-control guarantee.

%% file: sections/method.tex
\section{Method}\label{sec:method}

\Method separates tissue detection into initial segmentation, selective correction, and review. Fig.~\ref{fig:pipeline} illustrates how these stages interact: a thumbnail detector provides an initial mask, an audit network identifies regions likely to contain errors, and a frozen pathology foundation model refines only the highest-risk regions. Consensus fusion constrains the corrections, while a slide-level review flag identifies potentially unreliable outputs. Algorithm~\ref{alg:scout} summarizes the complete inference procedure. The output mask has a resolution of \MaskMpp{}\,\textmu m/px, and spatial parameters are specified in physical units.

\begin{algorithm}[t]
\caption{\Method on one slide}\label{alg:scout}
\begin{algorithmic}[1]
\Require WSI; audit head $A$; frozen FM $F$ with decoder $D$; thresholds $t$, $\tau$
\Ensure tissue mask $M$ at \MaskMpp{}\,\textmu m/px; review flag
\State $I \gets$ thumbnail at \MaskMpp{}\,\textmu m/px; $P \gets$ unscanned area; set $I$ to white on $P$
\State $B \gets$ OD threshold of $I$, closing and hole filling \Comment{base mask}\label{alg:base}
\State $(p_{\mathrm{FN}}, p_{\mathrm{FP}}) \gets A(I, \mathrm{OD}(I), B, d_B)$ \Comment{audit head}\label{alg:audit}
\State $\mathcal{C} \gets$ \WinUm{}\,\textmu m windows within \NearMm{}\,mm of $B$
\State $\mathcal{S} \gets$ the \BudgetPct\,\% of $\mathcal{C}$ with the largest $r_j=\sum_{x\in j}(p_{\mathrm{FN}}+p_{\mathrm{FP}})$\label{alg:selection}
\State $M \gets B$
\For{$j \in \mathcal{S}$}
\State $q \gets D\big(F(\text{window } j \text{ at } \FmMpp{}\,\text{\textmu m/px}), I_j\big)$\label{alg:fm}
\State $M_j \gets (q \ge t) \wedge (q \ge \FusionCut \vee B_j)$ \Comment{consensus}\label{alg:fusion}
\EndFor
\State $M \gets M \setminus P$; remove components below \MinCompMm{}\,mm$^2$\label{alg:cleanup}
\State $s \gets$ logistic regression on GT-free slide features\label{alg:review}
\State \Return $M$, flag $s > \tau$
\end{algorithmic}
\end{algorithm}

\subsection{Risk-guided tissue correction}
The first stage of Fig.~\ref{fig:pipeline} establishes both an initial tissue mask and a spatial estimate of its errors. The base detector thresholds the sum of the thumbnail's three optical-density (OD) channels at \BaseOd{}, applies morphological closing at \BaseCloseUm{}\,\textmu m, and fills holes below \BaseHoleMm{}\,mm. These parameters are selected on the development set. Large opaque-black regions introduced by some scanners outside the scanned area are set to white before detection and excluded from the final mask.

The audit head is a U-Net with a MobileNetV3-small encoder and \AuditParamsM{}\,M parameters. It receives the RGB thumbnail, its OD, the base mask, and the signed distance to the base-mask boundary. Rather than predicting a replacement segmentation directly, it estimates where the base detector misses tissue or incorrectly includes background. These two error-probability maps are learned from the base detector's errors on development slides. Lines~\ref{alg:base} and~\ref{alg:audit} of Algorithm~\ref{alg:scout} describe the initial mask and its audit, respectively.

The audit maps determine where additional computation is allocated. The thumbnail is partitioned into \WinUm{}\,\textmu m windows, with candidates restricted to those within \NearMm{}\,mm of the base mask. Each candidate is ranked by the sum of its predicted false-negative and false-positive probabilities over all pixels. As specified in line~\ref{alg:selection} of Algorithm~\ref{alg:scout}, only the highest-risk \BudgetPct\,\% of candidates are escalated. This selection links the audit output in the first stage of Fig.~\ref{fig:pipeline} to foundation-model inference in the second stage.

Each selected window is sampled at \FmMpp{}\,\textmu m/px with dimensions of $\FmInputPx\times\FmInputPx$ pixels and encoded by a frozen Virchow2 model~\cite{zimmermann2024virchow2}. A lightweight decoder maps the resulting representations to tissue probabilities on the thumbnail grid. Specifically, it projects the $16\times16$ patch-token grid to 128 channels through a $1\times1$ convolution, upsamples it to $\WinPx\times\WinPx$ pixels, concatenates the RGB thumbnail window, and applies three convolutions. Line~\ref{alg:fm} of Algorithm~\ref{alg:scout} describes this local refinement.

The decoder is trained on development windows where classical detectors disagree, using binary cross-entropy against the ground-truth (GT) masks and excluding the boundary ignore band defined below. It uses AdamW (learning rate \FmLr, weight decay \FmWd), batch size \FmBatch, flips and $90^\circ$ rotations, and at most \FmEpochs{} epochs, retaining the checkpoint with the best validation performance. Its decision threshold $t=\FmHeadThr$ is the lowest threshold for which false-positive tissue on the validation fold does not exceed \FmFpTarget\,\% of the GT area. The audit head is trained in the same way on the base detector's errors, from random $\AuditCropPx\times\AuditCropPx$ crops centred on candidate pixels (AdamW, learning rate \AuditLr, weight decay \AuditWd, one-cycle schedule, batch size \AuditBatch, at most \AuditEpochs{} epochs); out-of-fold heads are used for all development analyses.

\subsection{Consensus fusion}
The third stage of Fig.~\ref{fig:pipeline} combines local corrections with the initial mask. Outside the selected windows, the base prediction is retained. Within a selected window, retaining a base-positive pixel requires the foundation-model probability to reach $t$, whereas adding tissue to a base-negative pixel requires the higher probability \FusionCut. This asymmetric rule allows recovery of missed tissue while limiting the misclassification of unstained glass and debris. Line~\ref{alg:fusion} of Algorithm~\ref{alg:scout} gives the corresponding Boolean operation. Finally, components smaller than \MinCompMm{}\,mm$^2$ are removed, and unscanned regions are excluded, as specified in line~\ref{alg:cleanup}.

\subsection{Failure definition and calibrated review}\label{sec:failure}
The review mechanism requires an explicit definition of a mask that needs correction. Before evaluation, GT holes narrower than \HoleMm{}\,mm are filled, and a boundary band of $\pm$\BandUm{}\,\textmu m is ignored. A slide fails if any of four conditions holds: missed tissue exceeds \FailEps\,\% of the GT area; a fragment of at least \FragMinMm{}\,mm$^2$ has recall below \FragRecall{}; false-positive tissue exceeds \FailFp\,\% of the GT area; or precision falls below \FailPrec{}. These thresholds define an operational correction criterion and have not been validated against clinical consequences.

Evaluation and calibration use case groups to account for related slides. A group corresponds to a patient or case when the source identifies one (TCGA, MONKEY, NADT, ACROBAT), to a tissue-microarray block for DLBCL-Morphology, and otherwise to an individual slide. Excluding the per-slide groups leaves \SingNGroups{} test groups and the same comparison (\Method \SingFailTrace\,\%, AtlasPatch \SingFailAtlaspatch\,\%, U-Net \SingFailUnet\,\%). A group fails if any of its slides fails.

Automated correction is complemented by the review flag shown at the end of Fig.~\ref{fig:pipeline}. A logistic regression fitted on development slides assigns each slide a score using features available without GT: estimated missed and false-positive tissue from the audit and decoder outputs, the escalated fraction, foundation-model uncertainty, changes to the base mask, component counts, and the recorded stain family. A slide is flagged when its score exceeds a calibrated threshold $\tau$, as summarized in line~\ref{alg:review} and the return step of Algorithm~\ref{alg:scout}.

Calibration controls silent failures, defined as case groups containing at least one failed slide that is not flagged. Let $R(\tau)$ denote the probability of this event for a randomly drawn case group. Increasing the threshold reduces the number of review flags and can therefore increase this risk. Learn-then-Test fixed-sequence testing~\cite{angelopoulos2025ltt} selects a threshold using the calibration groups. Starting from a threshold that flags every slide, it tests progressively larger thresholds against a target risk $\alpha$. Each test uses the lower-tail binomial probability of observing no more than the measured number of silent calibration-group failures, assuming failure probability $\alpha$. The procedure stops at the first probability exceeding \CertDelta{} and retains the preceding threshold; if no threshold passes, all slides are flagged.

For case groups drawn independently from the population represented by the benchmark's sources, the selected threshold $\hat\tau$ satisfies
\begin{equation}
\Pr\!\left\{R(\hat\tau)\le\alpha\right\}\ge 1-\CertDelta.
\label{eq:review-guarantee}
\end{equation}
This guarantee concerns the marginal proportion of all case groups containing an unflagged failure. It does not bound the conditional failure rate among accepted slides or provide a separate guarantee for each source.

%% file: sections_cmpb/experiments.tex
\section{Experimental setup}\label{sec:exp}

The evaluation examines whether selective correction reduces operational tissue-detection failures, whether the improvement persists on sources excluded from task-specific training, and whether the resulting masks preserve downstream classification performance. The main comparison uses a multi-stain benchmark, complemented by source-held-out experiments and external evaluations. Table~\ref{tab:main} reports the primary comparison, Table~\ref{tab:loso} addresses source generalization, and Table~\ref{tab:ablation} isolates the contributions of individual components.

\begin{figure*}[!t]\centering
\includegraphics[width=\textwidth]{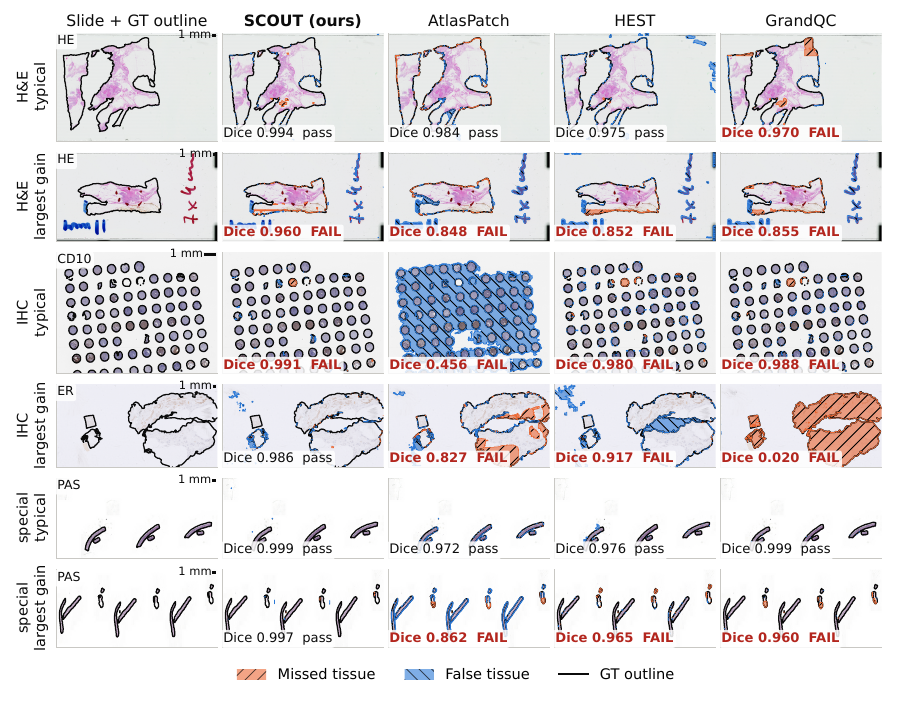}
\caption{Prespecified test examples by stain: \emph{typical}, median \Method Dice; \emph{largest gain}, greatest Dice improvement over the best comparator. Comparators are DEV-tuned. Hatching denotes missed ($/$) and false ($\backslash$) tissue. Creative Commons images, cropped around tissue.}\label{fig:qualitative}
\end{figure*}

\subsection{Datasets and partitioning}
Tissues with Issues (TwI)~\cite{polejowska2026twi} provides manually corrected tissue masks from multiple sources, organs, and stains. Slides are partitioned by the case groups defined in Section~\ref{sec:failure}, with stratification by source. The development set contains \NumDevSlides{} slides in \NumDevGroups{} groups, the calibration set contains \NumCalSlides{} slides in \NumCalGroups{} groups, and the test set contains \NumTestSlides{} slides in \NumTestGroups{} groups. Development data are used to fit model components and select segmentation thresholds and comparator thresholds and post-processing. Calibration data are reserved for selecting the review threshold $\tau$, while the test set evaluates the final system and its ablations.

An additional external assessment uses the \NumBandi{} manually labelled slides of B\'andi et al.~\cite{bandi2019,bandi2019data}. Three B\'andi slides used for tuning are reported separately. The downstream evaluation uses CAMELYON17, as described below.

\subsection{Comparators and evaluation protocol}
We compare \Method with five publicly available detectors: GrandQC~\cite{weng2024grandqc}, HEST~\cite{jaume2024hest}, AtlasPatch~\cite{alagha2026atlaspatch}, the nnU-Net of Helgesen et al.~\cite{helgesen2026}, and HistoQC~\cite{janowczyk2019histoqc}. Each is evaluated in its released configuration and after optimization on the development set, denoted \emph{DEV-tuned}. For the four detectors that output a tissue probability, the latter selects jointly the probability threshold (\CNineNT{} values from \CNineTMin{} to \CNineTMax) and one of \NumTuneCombos{} combinations of morphology, dilation or erosion, and small-object and hole filtering, by the lowest development failure rate under the operational criterion; for HistoQC only the post-processing is selected. This adapts outputs to the benchmark's annotation convention while retaining the pretrained representations. A first evaluation tuned the post-processing at the released threshold only; the threshold search was specified before its results were known, replaces it in all tables, and the earlier results are reported where they differ. HistoQC intentionally excludes fat and blurred regions, whereas GrandQC was developed for H\&E; these differences are relevant when evaluating against total-tissue IHC references.

The tuned OD threshold and CLAM~\cite{lu2021clam} provide threshold-based baselines. To distinguish the benefit of the cascade from adaptation to the development annotations, we also train a U-Net with an ImageNet-initialized EfficientNet-B0 encoder on the same development masks at thumbnail resolution. Training uses three seeds, early stopping on one development fold, and a threshold selected to maximize the development pass rate. A variant adds HED stain augmentation~\cite{tellez2019augmentation} (per-channel factor $U(0.95, 1.05)$, shift $U(-0.05, 0.05)$ in optical density) and $\pm10$\,\% brightness and contrast jitter.

The primary endpoint is the test-set case-group failure rate under the correction criterion in Section~\ref{sec:failure}. Failure-rate uncertainty is quantified using Clopper--Pearson intervals~\cite{clopper1934}. Between-method differences are estimated by paired bootstrap resampling of case groups with \NBoot{} resamples; hypothesis tests are two-sided, with Holm correction for multiple comparisons~\cite{holm1979}.

Complementary measures describe the nature and spatial extent of segmentation errors: missed and false-positive tissue, Dice overlap, the 95th-percentile Hausdorff distance (HD95), and the boundary F-score at \BandUm{}\,\textmu m (BF16). Table~\ref{tab:stain} reports stain-specific failure rates, while Fig.~\ref{fig:qualitative} illustrates segmentation behavior using the prespecified selection rules detailed in its caption.

To assess generalization beyond sources represented in task-specific training, we exclude each of the \NumLosoSources{} sources with at least nine test groups in turn. For each held-out source, we refit the base threshold, audit and foundation-model heads, U-Net, CNN tier, and comparator thresholds and post-processing without that source, then evaluate on its test groups. Table~\ref{tab:loso} and Fig.~\ref{fig:results}b summarize this evaluation. Because one source could dominate the pooled result, we also repeat the pooled comparison leaving out each held-out source in turn. The exclusion applies to task-specific fitting and cannot establish that a source was absent from foundation-model pretraining; Virchow2, for example, was pretrained on MSKCC slides~\cite{zimmermann2024virchow2}.

\subsection{Downstream evaluation}
The downstream experiment tests whether mask refinement preserves diagnostically relevant input for slide classification. We use the \SixNSlides{} CAMELYON17 training slides~\cite{bandi2019camelyon17}, comprising lymph nodes from five centres, none used for development. Each detector's mask selects \SixPatchUm\,\textmu m patches for an attention-based multiple-instance classifier~\cite{ilse2018abmil} using frozen H0-mini features~\cite{filiot2025h0mini}. The classifier distinguishes metastatic from negative slides using leave-one-centre-out folds and five seeds. On the \SixNAnnotated{} slides with lesion polygons, we additionally measure annotated lesion area excluded by each mask. Unscanned black padding is removed from every mask. AtlasPatch was trained on CAMELYON17, making this evaluation set in-domain for AtlasPatch alone.

%% file: sections_cmpb/software.tex
\section{Software}\label{sec:software}
Because the Virchow2 licence prohibits redistribution of models trained on its outputs, the released software runs \Method-open, the same cascade with the Apache-licensed H-optimus-0 foundation model~\cite{saillard2024hoptimus0}. On the test set \Method-open fails on \SevenFailHopt\,\% of the case groups, against \FailTrace\,\% for \Method (Table~\ref{tab:ablation}).

\emph{Package.} \texttt{scout-tissue} is a Python package (Apache-2.0; Python $\ge$ 3.10) distributed on PyPI. It contains only the inference path of Algorithm~\ref{alg:scout}: reading the slide at a fixed physical resolution with OpenSlide, the base detector, the audit head and risk gate, the FM tier with its decoder, consensus fusion and export. The trained heads are downloaded from Hugging Face at a pinned revision and checked against their SHA-256; H-optimus-0 is obtained from its provider. Only pyramid levels are read, so label and macro images, which may carry patient identifiers, are never opened, and slides without usable resolution metadata are rejected rather than read at a guessed scale.

\emph{Interfaces.} A command-line tool processes one or many slides with the models loaded once (\texttt{scout-tissue predict slide.svs --out masks/}). The Python interface returns the mask with its grid, so that it can be placed in any downstream pipeline:
\begin{verbatim}
from scout_tissue import Scout
scout = Scout.from_pretrained(device="cuda")
res = scout.predict("slide.svs")
res.mask, res.grid, res.to_geojson()
\end{verbatim}
Masks are written as compressed NumPy arrays with the slide-to-mask transform (origin and scale in level-0 pixels), as PNG, or as GeoJSON polygons in level-0 coordinates that QuPath imports as annotations. A browser demo (\texttt{scout-tissue demo}) shows the thumbnail, the mask, the escalated windows and the audit risk map.

\emph{Verification.} The package was checked against the reference implementation that produced the results of this paper: on \PkgParitySlides{} slides (\PkgParityTwi{} test slides of all stain families and \PkgParityBandi{} B\'andi slides) the masks of the installed package are identical to the reference ones (\PkgParityMaxDiffPx{} differing pixels), with identical grid metadata. Unit tests on synthetic data cover fusion, quantization, coordinate conversion and export. Inference time is dominated by the FM tier (\RtTraceFmShare\,\% of the total for \Method), so it grows with the number of windows sent to it; a GPU is recommended.

\emph{Availability.} Source code: \url{https://github.com/CosmoIknosLab/scout-tissue-detection}; package: \url{https://pypi.org/project/scout-tissue/}; weights (CC BY-NC 4.0): \url{https://huggingface.co/Cristiandieti/scout-tissue-detection}; archived release: \href{https://doi.org/10.5281/zenodo.23664072}{doi:10.5281/zenodo.23664072}.

%% file: sections_cmpb/results.tex
\section{Results}\label{sec:results}

Released detectors need correction far more often than their mean overlap suggests, even with their threshold and post-processing tuned on the development set. On sources represented in training, a U-Net trained on the benchmark's masks removes most of these failures, as does \Method. On held-out sources, \Method and the re-tuned HEST fail equally often overall but on different sources. With a single configuration, \Method has the lowest failure rate, alone or tied, on represented sources, on held-out sources and on the external slides, although not always significantly lower. The analyses below then relate \Method's behaviour to inference allocation, consensus fusion, review workload and downstream performance.

\input{generated/tab_main}

\subsection{Released detectors and adaptation to the annotation convention}
Table~\ref{tab:main} shows that \Method reduces case-group failures to \FailTrace\,\% from \FailHestDevtunedT\,\% for DEV-tuned HEST, the strongest released comparator. The paired difference is \DiffTraceHestDevtunedT{} percentage points (pp), with a 95\,\% confidence interval of [\DiffTraceHestDevtunedTLo, \DiffTraceHestDevtunedTHi] ($p=\PTraceHestDevtunedT$); against the other released detectors it is at least \DiffTraceAtlaspatchDevtunedTAbs{} pp. Tuning the threshold matters: the selected thresholds lie below the released 0.5 (\CNineTHest{} for HEST, \CNineTAtlaspatch{} for AtlasPatch), and with post-processing tuned alone HEST fails on \FailHestDevtuned\,\% of the groups. Compared with HEST, \Method has fewer slide failures from missed tissue and from false tissue (m/f/b column) and the best boundary scores. Mean Dice alone obscures this distinction: HEST achieves \DiceHestDevtunedT{} despite failing on \FailHestDevtunedT\,\% of groups. On the calibration groups \Method fails on \CalFailTrace\,\% and HEST on \CalFailHestDevtunedT\,\%. Against the comparators tuned at their released threshold, the ranking also persisted under alternative missed-tissue limits, a miss-only criterion, the raw annotation convention and the exclusion of suspect GT regions (components of at least 0.1\,mm$^2$ that every released detector contradicts); these variants were not repeated for the threshold-tuned comparators.

Across \NumSeedsTrace{} training seeds, \Method fails on \SeedFailMean{} $\pm$ \SeedFailSd\,\% of groups, compared with \UnetFailMean{} $\pm$ \UnetFailSd\,\% for the U-Net. Thus, on represented sources, adaptation to the benchmark's annotation convention largely closes the gap between a conventional segmenter and the cascade.

\input{generated/tab_stain}

The stain-specific results in Table~\ref{tab:stain} identify IHC as the principal challenge for released detectors. \Method fails on \StainIhcFailTrace\,\% of IHC groups, compared with at least \StainIhcFailReleasedMin\,\% for the released comparators, and has the lowest H\&E failure rate. The IHC rows of Fig.~\ref{fig:qualitative} illustrate recovery of pale tissue on near-white backgrounds. The U-Net performs similarly on represented sources; the distinction becomes clearer under source-held-out evaluation.

\begin{figure*}[!t]\centering
\includegraphics[width=\textwidth]{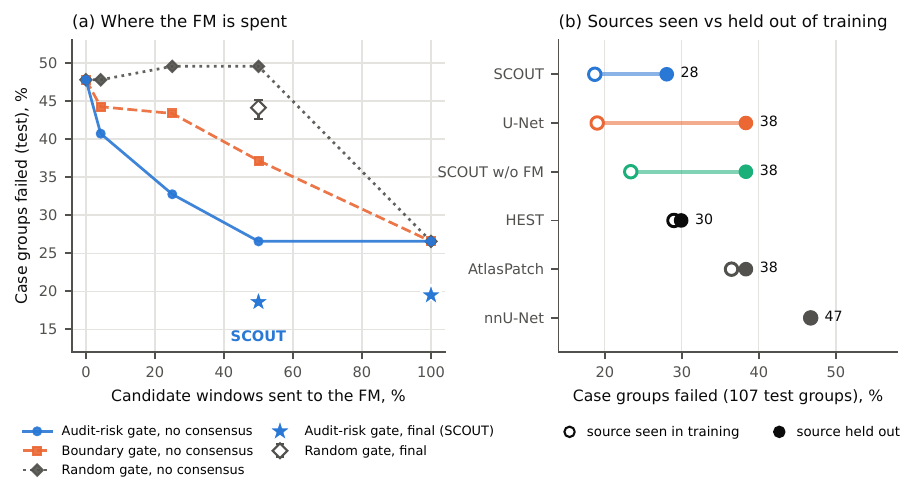}
\caption{(a) Failure rate versus FM budget: lines omit consensus, stars include it; random-gate stars show mean and range over \RbRandomN{} orderings. (b) Source-present (open) versus source-held-out (filled) failure rates on \LosoNGroups{} groups; released detectors with threshold and post-processing tuned.}\label{fig:results}
\end{figure*}

\input{generated/tab_loso}

Table~\ref{tab:loso} and Fig.~\ref{fig:results}b show a smaller decline for \Method than for the U-Net when sources are excluded from training. Its pooled failure rate is \LosoFailTrace\,\%, compared with \LosoFailUnet\,\% for the U-Net and \LosoFailCnnTier\,\% for the cascade with a CNN replacing the foundation model. The difference relative to the U-Net persists across three training seeds, with a paired difference of \LosoSeedDiffUnet{} pp and a 95\,\% confidence interval of [\LosoSeedDiffUnetLo, \LosoSeedDiffUnetHi], and is not eliminated by stain augmentation (\AugSixBLoso\,\% for the augmented U-Net; \AugSixBLosoDiff{} pp [\AugSixBLosoDiffLo, \AugSixBLosoDiffHi]). The re-tuned HEST, whose threshold and post-processing were re-selected without the held-out source like every other fitted part, fails on \LosoFailHestT\,\%, which is statistically indistinguishable from \Method (\LosoDiffHestT{} pp [\LosoDiffHestTLo, \LosoDiffHestTHi], $p=\LosoPHestT$). The other released detectors fail more often (AtlasPatch \LosoFailAtlaspatchT\,\%, \LosoDiffAtlaspatchT{} pp [\LosoDiffAtlaspatchTLo, \LosoDiffAtlaspatchTHi]). No method dominates every source; \Method has the lowest macro-average failure rate (\LosoMacroTrace\,\% against \LosoMacroHestT\,\% for HEST) and shares the lowest worst-source rate with HEST. HEST retains the highest pooled Dice despite more case-group failures.

The pooled comparisons depend on one source. On ACROBAT, HEST fails on \LosoAcrobatFailHestT\,\% of the groups and \Method on \LosoAcrobatFailTrace\,\%; leaving ACROBAT out, \Method fails on \LosoNoAcrobatFailTrace\,\% of the remaining \LosoNoAcrobatNGroups{} groups against \LosoNoAcrobatFailHestT\,\% for HEST (\LosoNoAcrobatDiffHestT{} pp [\LosoNoAcrobatDiffHestTLo, \LosoNoAcrobatDiffHestTHi]) and \LosoNoAcrobatFailUnet\,\% for the U-Net, a difference of \LosoNoAcrobatDiffUnet{} pp [\LosoNoAcrobatDiffUnetLo, \LosoNoAcrobatDiffUnetHi] that is not statistically significant ($p=\LosoNoAcrobatPUnet$). Against the cascade without the FM the difference remains significant (\LosoNoAcrobatDiffCnn{} pp [\LosoNoAcrobatDiffCnnLo, \LosoNoAcrobatDiffCnnHi]). Leaving out any other source, the difference to the U-Net ranges from \LosoNoOtherDiffUnetMin{} to \LosoNoOtherDiffUnetMax{} pp, with every upper confidence bound at or below \LosoNoOtherDiffUnetHiMax{} pp, and the difference to HEST from \LosoNoOtherDiffHestTMin{} to $+$\LosoNoOtherDiffHestTMax{} pp, none significant. On held-out sources, therefore, \Method's advantage over the U-Net and its parity with the re-tuned HEST both rest substantially on ACROBAT, whereas the advantage over the same cascade without an FM holds without it.

\input{generated/tab_ablation}

\subsection{Where the foundation model is spent}
Table~\ref{tab:ablation} and Fig.~\ref{fig:results}a show that inference allocation, rather than processing volume alone, determines the benefit of the foundation model. Random selection provides limited improvement over the base detector, whereas audit-guided selection substantially reduces failures. Processing every candidate yields no further reduction, so the gate halves the number of foundation-model windows without increasing the observed failure rate.

Consensus fusion limits the false-positive tissue introduced by unconstrained foundation-model predictions, and small-component removal improves HD95. Candidate coverage is nearly complete: at most \CandOutsideMaxPct\,\% of GT tissue lies outside it. Replacing predictions in selected windows with GT leaves \OracleSelTest\,\% of groups failing, suggesting that most residual failures arise within those windows. Adding votes from six detectors---grayscale and saturation Otsu, CLAM, EntropyMasker~\cite{song2023entropy}, Double-Pass~\cite{ceachi2025doublepass}, and GrandQC---does not improve performance. Alternative frozen foundation models yield similar results, whereas the CNN tier is less robust, particularly on unseen sources.

\subsection{Review, external and downstream evaluation}
At a target marginal risk of \CertAlpha\,\%, the review flag identifies \FlagSensTen\,\% of failed groups while flagging \CertTraceFlagTen\,\% of test slides. Unflagged failures remain in \CertTraceSilentTen{} of \NumTestGroups{} groups, compared with \RbRandFlagSilent{} on average for random flagging at the same rate. Reducing the target to \CertAlphaFive\,\% leaves \CertTraceSilentFive{} unflagged group failures but requires review of \FlagGroupsFive\,\% of groups, illustrating the trade-off between review workload and silent failures. None of the comparators provides a calibrated signal of this kind.

On the \NumBandiStrict{} B\'andi slides excluded from tuning, \Method fails on \BandiFailTraceStrict, compared with \BandiStrictUnet{} for the U-Net and, for the DEV-tuned detectors, \BandiStrictGrandqcT{} for GrandQC, \BandiStrictHelgesenT{} for nnU-Net, \BandiStrictAtlaspatchT{} for AtlasPatch and \BandiStrictHestT{} for HEST. On all \NumBandi{} slides the threshold that makes HEST competitive on held-out sources fails \BandiFailHestT{} of them, against \BandiFailTrace{} for \Method, mostly by missing small tissue fragments.

On CAMELYON17, \Method masks yield a classification area under the receiver operating characteristic curve of \SixAucTrace{}. Under the predeclared margin of \SixMargin, non-inferiority is established relative to AtlasPatch and GrandQC, with lower 95\,\% difference bounds of \SixAucDiffAtlaspatchLo{} and \SixAucDiffGrandqcLo{}, respectively; it is not established relative to HEST. \Method excludes \SixLostTrace\,\% of annotated lesion area, compared with \SixLostGrandqc\,\% for GrandQC and \SixLostClam\,\% for CLAM. This experiment used the comparators with post-processing tuned at their released threshold.

Starting from the raw slide on an RTX~3090, median runtime is \RtTraceGpu\,s for \Method, \RtHestGpu\,s for HEST, and \RtGrandqcGpu\,s for GrandQC; foundation-model inference accounts for \RtTraceFmShare\,\% of \Method's runtime.

%% file: generated/tab_main.tex
\begin{table*}[t]
\centering
\footnotesize
\setlength{\tabcolsep}{3.5pt}
\caption{TwI test results. Brackets: 95\% confidence intervals for failure rates (Clopper--Pearson) and paired differences (group bootstrap). Missed/false: mean per-slide fraction of GT area. m/f/b: missed-only/false-only/both slide failures. Released detectors: threshold and post-processing tuned on DEV (HistoQC: post-processing only). U-Net and SCOUT: seed 0; nnU-Net:~\cite{helgesen2026}. Best values in bold.}
\label{tab:main}
\begin{adjustbox}{max width=\textwidth}
\begin{tabular}{lrrrrrrrr}
\toprule
Method & Groups failed, \% & SCOUT $-$ method, pp & Missed, \% & False, \% & Dice & HD95, $\mu$m & BF16 & m/f/b \\
\midrule
\textbf{SCOUT (ours)} & \textbf{18.6 [11.9, 27.0]} & reference & \textbf{1.0} & 1.4 & \textbf{0.988} & \textbf{466} & \textbf{0.68} & 25/3/0 \\
U-Net (DEV-trained) & 19.5 [12.6, 28.0] & $-0.9$ [-7.1, +4.4] & 1.1 & 1.3 & \textbf{0.988} & -- & -- & 24/1/2 \\
HEST (DEV-tuned) & 29.2 [21.0, 38.5] & $-10.6$ [-20.4, -0.9] & 1.1 & 3.3 & 0.979 & 496 & 0.57 & 31/4/4 \\
AtlasPatch (DEV-tuned) & 37.2 [28.3, 46.8] & $-18.6$ [-27.4, -9.7] & 2.4 & 9.5 & 0.952 & 681 & 0.33 & 29/18/5 \\
nnU-Net (DEV-tuned) & 46.9 [37.5, 56.5] & $-28.3$ [-37.2, -19.5] & 2.5 & 15.5 & 0.926 & 1199 & 0.26 & 30/27/19 \\
Tuned threshold & 47.8 [38.3, 57.4] & $-29.2$ [-38.1, -20.4] & 12.7 & 10.1 & 0.877 & 4018 & 0.44 & 44/17/23 \\
GrandQC (DEV-tuned) & 52.2 [42.6, 61.7] & $-33.6$ [-43.4, -23.9] & 30.9 & 1.1 & 0.712 & 1199 & 0.49 & 74/2/9 \\
CLAM (default) & 82.3 [74.0, 88.8] & $-63.7$ [-72.6, -54.9] & 46.6 & 5.8 & 0.565 & 3334 & 0.17 & 114/5/12 \\
HistoQC (DEV-tuned) & 94.7 [88.8, 98.0] & $-76.1$ [-84.1, -68.1] & 60.2 & \textbf{0.4} & 0.474 & 4951 & 0.03 & 146/0/1 \\
\bottomrule
\end{tabular}
\end{adjustbox}
\end{table*}

%% file: generated/tab_stain.tex
\begin{table}[t]
\centering
\footnotesize
\setlength{\tabcolsep}{3.5pt}
\caption{Failure rates (\%) by stain; group counts in headers. Multi-stain groups enter each relevant column. Released detectors: threshold and post-processing DEV-tuned; U-Net: seed 0. Best values in bold.}
\label{tab:stain}
\begin{adjustbox}{max width=\columnwidth}
\begin{tabular}{lrrr}
\toprule
Method & H\&E (78) & IHC (50) & Special (17) \\
\midrule
\textbf{SCOUT (ours)} & \textbf{16.7} & \textbf{22.0} & 5.9 \\
U-Net & 17.9 & 26.0 & \textbf{0.0} \\
AtlasPatch & 17.9 & 62.0 & 17.6 \\
HEST & 21.8 & 28.0 & 29.4 \\
nnU-Net & 34.6 & 72.0 & 29.4 \\
GrandQC & 33.3 & 100.0 & 11.8 \\
Tuned threshold & 48.7 & 82.0 & 11.8 \\
\bottomrule
\end{tabular}
\end{adjustbox}
\end{table}

%% file: generated/tab_loso.tex
\begin{table}[t]
\centering
\footnotesize
\setlength{\tabcolsep}{3.5pt}
\caption{Source-held-out failure rates (top) and mean Dice (bottom). U-Net: three-seed mean. CNN tier replaces the FM. Released detectors: threshold and post-processing re-selected without the held-out source. Macro: equal source weights; worst: least favorable source. Best values in bold.}
\label{tab:loso}
\begin{adjustbox}{max width=\columnwidth}
\begin{tabular}{lrrrrrrr}
\toprule
Held out & Groups & SCOUT & U-Net & CNN tier & nnU-Net & AtlasPatch & HEST \\
\midrule
\multicolumn{8}{l}{\emph{Groups failed, \%}} \\
ACROBAT & 30 & 66.7 & 87.8 & 86.7 & 86.7 & 80.0 & \textbf{26.7} \\
TCGA & 43 & \textbf{14.0} & 14.7 & 18.6 & 25.6 & \textbf{14.0} & 27.9 \\
MONKEY & 16 & \textbf{0.0} & 4.2 & \textbf{0.0} & 25.0 & 31.2 & 31.2 \\
MOSTWIEDZY & 9 & 22.2 & 48.1 & 44.4 & 44.4 & 22.2 & \textbf{11.1} \\
MSKCC & 9 & \textbf{22.2} & 37.0 & 33.3 & 55.6 & 44.4 & 66.7 \\
\cmidrule(l){1-8}
Pooled & 107 & \textbf{28.0} & 38.3 & 38.3 & 46.7 & 38.3 & 29.9 \\
Macro average & -- & \textbf{25.0} & 38.4 & 36.6 & 47.4 & 38.4 & 32.7 \\
Worst source & -- & \textbf{66.7} & 87.8 & 86.7 & 86.7 & 80.0 & \textbf{66.7} \\
\midrule
\multicolumn{8}{l}{\emph{Mean Dice}} \\
ACROBAT & 30 & 0.962 & 0.896 & 0.908 & 0.897 & 0.941 & \textbf{0.974} \\
TCGA & 43 & \textbf{0.988} & 0.983 & 0.983 & 0.975 & 0.987 & 0.984 \\
MONKEY & 16 & \textbf{0.997} & 0.992 & 0.992 & 0.972 & 0.955 & 0.990 \\
MOSTWIEDZY & 9 & 0.984 & 0.939 & 0.959 & 0.910 & 0.972 & \textbf{0.989} \\
MSKCC & 9 & \textbf{0.988} & 0.981 & 0.983 & 0.932 & 0.933 & 0.965 \\
\cmidrule(l){1-8}
Pooled & 107 & 0.977 & 0.943 & 0.950 & 0.933 & 0.959 & \textbf{0.979} \\
Macro average & -- & \textbf{0.984} & 0.958 & 0.965 & 0.938 & 0.958 & 0.980 \\
Worst source & -- & 0.962 & 0.896 & 0.908 & 0.897 & 0.933 & \textbf{0.965} \\
\bottomrule
\end{tabular}
\end{adjustbox}
\end{table}

%% file: generated/tab_ablation.tex
\begin{table}[t]
\centering
\footnotesize
\setlength{\tabcolsep}{3.5pt}
\caption{TwI ablation. Failed/Unseen: case-group failure rates (\%); Unseen uses source-held-out evaluation. Missed/False: \% of GT area; HD95: median in $\mu$m. Detector votes augment audit inputs and candidates. Best values in bold.}
\label{tab:ablation}
\begin{adjustbox}{max width=\columnwidth}
\begin{tabular}{lrrrrr}
\toprule
Variant & Failed & Missed & False & HD95 & Unseen \\
\midrule
Base threshold only & 47.8 & 12.7 & 10.1 & 4018 & 46.7 \\
+ FM on random 50\% & 49.6 & 6.4 & 9.5 & -- & -- \\
+ FM on riskiest 50\% & 26.5 & \textbf{0.6} & 7.9 & 2649 & -- \\
\quad (FM on all candidates) & 26.5 & \textbf{0.6} & 8.9 & -- & -- \\
+ consensus fusion & \textbf{18.6} & 1.0 & 1.6 & 1881 & -- \\
+ small-part filter & \textbf{18.6} & 1.0 & \textbf{1.4} & 463 & -- \\
\textbf{+ padding rule (SCOUT)} & \textbf{18.6} & 1.0 & \textbf{1.4} & 466 & \textbf{28.0} \\
\midrule
\quad + vote of six detectors & 19.5 & 1.0 & \textbf{1.4} & 471 & \textbf{28.0} \\
\quad FM: H-optimus-0 & 19.5 & 1.2 & \textbf{1.4} & \textbf{386} & -- \\
\quad FM: UNI2-h & \textbf{18.6} & 1.2 & 2.0 & 1218 & -- \\
\quad FM: H0-mini & 23.0 & 1.2 & 2.9 & 1275 & -- \\
\quad no FM: CNN tier & 23.0 & 1.3 & 2.5 & 1216 & 38.3 \\
\bottomrule
\end{tabular}
\end{adjustbox}
\end{table}

%% file: sections_cmpb/discussion.tex
\section{Discussion}\label{sec:discussion}
Measured by how often a mask needs correction, tissue detection is not solved: released detectors fail on a large share of case groups even after their threshold and post-processing are tuned to the benchmark, almost always by missing tissue that is pale, sparse or stained with something other than H\&E, and their mean Dice hides it. The first practical consequence is that a detector should be adapted to the annotation convention of its target data: on sources represented in training, a plain U-Net trained on the benchmark's masks removes most failures and is statistically indistinguishable from \Method. The second is that no adapted method is uniformly best when the source changes. On held-out sources the U-Net and the same cascade without an FM degrade more than \Method, while the re-tuned HEST fails as often as \Method overall but on different sources: it is better on ACROBAT and worse on the others, and both the gap to the U-Net and the parity with HEST rest substantially on ACROBAT. Only the gap to the cascade without an FM holds for every choice of held-out sources, so a frozen FM appears to add knowledge of tissue that the development masks do not contain; \Method uses it where the audit expects errors, at half the cost of applying it everywhere and without letting it invent tissue.

What distinguishes \Method is consistency rather than a large margin in any single setting. Each alternative falls behind it in at least one setting: the U-Net on held-out sources, and HEST on the represented sources and on the external slides, where the low threshold that helps it on held-out sources leaves \BandiFailHestT{} of \NumBandi{} slides failing. \Method reaches the lowest failure rate, alone or tied, in all three settings with one configuration, and it is the only method that reports which slides need review with a calibrated risk.

These results suggest a simple rule for practice. With manual masks from the target sites, a small segmentation network trained on them is a strong and cheap choice. For new sites, scanners or stains, or for heterogeneous archives, a released detector needs at least its threshold tuned on annotated slides from a similar source, and the tuned threshold may not carry over to the next one; the FM-based cascade is a more consistent default, and its calibrated flag tells the user which slides to check. On H\&E lymph nodes, where detection is easy, any good mask gives a similar downstream classifier, so the benefit lies in the harder stains and sources.

\emph{Limitations.} Evaluation relies on one benchmark, a small external set, and case groups that do not always represent patients; the ablations also use the test set. The correction criterion is operational and has not been validated against clinical consequences; the sensitivity variants of the criterion were computed against the comparators tuned at their released threshold. The comparator threshold search was added after a first evaluation and pre-specified before its results were known; it makes HEST much stronger, which shows that comparisons with released detectors depend on how they are tuned. Without the test groups that are single slides of unidentified patients, \Method's advantage over HEST on the test set is no longer significant. Reference-mask errors lack expert adjudication. The generalization evidence against the U-Net and HEST depends substantially on ACROBAT, and overlap between the held-out sources and FM pretraining data cannot be excluded. The review guarantee is marginal over case groups drawn from the benchmark's sources and does not bound per-site or accepted-slide failure rates. Downstream evidence is limited to one classification task, run with the comparators at their released threshold, where non-inferiority to HEST is not established. \Method requires a GPU.

\section{Conclusions}
Released tissue detectors need manual correction far more often than their mean overlap suggests, particularly for pale and non-H\&E tissue, even after their threshold and post-processing are tuned. Training on masks in the target convention removes most failures on familiar sources, and on unfamiliar sources no adapted method is uniformly best. Spending a frozen foundation model only on the windows a small audit network considers risky gives the most consistent detection across familiar, held-out and external slides at bounded cost. \Method combines this cascade with a calibrated review flag and is released as a verified, pip-installable package.

%% file: sections/supplement_body.tex
This supplement gives the case-group definitions, results by source, a stain-augmented U-Net, the operating points of the review flag, the lowest-Dice examples, the quality of the reference masks and the runtime. Section, table and figure numbers prefixed with S refer to this supplement; all others refer to the main text.

\section{Case groups}
Slides of one case group never cross splits, and a group fails if any of its slides fails. The grouping rule depends on what each source identifies (Table~\ref{tab:groups}): patients for TCGA, MONKEY and NADT, cases for ACROBAT, tissue-microarray blocks for DLBCL-Morphology and specimens for the external slides of B\'andi et al. Sources without a patient identifier are grouped per slide, so slides of one patient could fall into different splits.
\input{generated/tab_groups}

To check that these per-slide groups do not drive the results, we repeated the main comparison without them, which leaves \SingNGroups{} test case groups. On this subset \Method fails on \SingFailTrace\,\%, DEV-tuned HEST on \SingFailHest\,\% (difference \SingDiffHest{} pp [\SingDiffHestLo, \SingDiffHestHi]), DEV-tuned AtlasPatch on \SingFailAtlaspatch\,\% (\SingDiffAtlaspatch{} pp [\SingDiffAtlaspatchLo, \SingDiffAtlaspatchHi]) and the U-Net (seed 0) on \SingFailUnet\,\% (\SingDiffUnet{} pp [\SingDiffUnetLo, \SingDiffUnetHi]). The ordering is unchanged, but the difference to HEST is no longer significant.

\section{Results by source}
Table~\ref{tab:source} breaks the test failures of Table~\MainTabMain{} of the main text down by source. \Method has the fewest failed groups, alone or tied, on every source except ACROBAT, where HEST fails on two groups fewer, and MOSTWIEDZY, where the U-Net and HEST fail on one group fewer. These are sources seen in training; the source-held-out results are in Table~\MainTabLoso{} of the main text.
\input{generated/tab_source}

\section{U-Net with stain augmentation}
To test whether the U-Net's loss on unseen sources is a lack of colour augmentation, we retrained it with HED stain augmentation~\cite{tellez2019augmentation} (per-channel factor $U(0.95, 1.05)$ and shift $U(-0.05, 0.05)$ in optical density) and $\pm10$\,\% brightness and contrast jitter, with three seeds and otherwise unchanged settings. It fails on \AugSixBFail\,\% of the test case groups (plain U-Net \UnetFailMean\,\%) and on \AugSixBLoso\,\% under source-held-out evaluation (plain U-Net \LosoFailUnet\,\%); \Method minus the augmented U-Net is \AugSixBLosoDiff{} pp [\AugSixBLosoDiffLo, \AugSixBLosoDiffHi]. A stronger variant, applying the same shift in the rescaled stain space of scikit-image ($\approx\pm0.69$ optical density), fails on \AugSixFail\,\% and \AugSixLoso\,\%, respectively. Augmentation therefore does not close the gap on unseen sources.

\section{Review flag}
Table~\ref{tab:flag} gives the operating points of the review flag. The threshold $\tau$ is chosen on the calibration groups and applied unchanged to the test groups. At $\alpha=\CertAlpha\,\%$ the flag sends \CertTraceFlagTen\,\% of the test slides to review and catches \FlagSensTen\,\% of the failed groups; at $\alpha=\CertAlphaFive\,\%$ it catches \FlagSensFive\,\% of them but flags \CertTraceFlagFive\,\% of the slides. On both sets the observed rate of silent groups stays below $\alpha$. The last column is the failure rate among the groups that are not flagged, which the bound does not control.
\input{generated/tab_flag}

\section{Lowest-Dice examples}
\begin{figure}[H]\centering
\includegraphics[width=\textwidth]{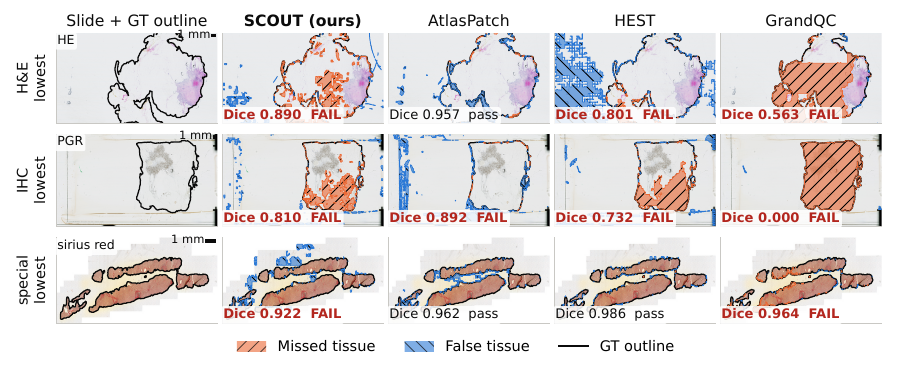}
\caption{Test slides with the lowest \Method Dice per stain family (Creative Commons licences), cropped around the tissue; comparators are DEV-tuned (threshold and post-processing). Hatching marks missed ($/$) and false ($\backslash$) tissue.}\label{fig:qualsupp}
\end{figure}

Fig.~\ref{fig:qualsupp} shows, for each stain family, the test slide on which \Method has its lowest Dice, with the same columns as Fig.~2 of the main text. These are \Method's worst cases and all three fail. On the IHC slide every method shown fails, and on the H\&E slide only AtlasPatch passes; on the special stain \Method adds false tissue next to the section, where AtlasPatch and HEST pass.

\section{Reference quality}
The reference masks contain errors. Table~\ref{tab:gt} reports two checks. \emph{Suspect regions} are components of at least 0.1\,mm$^2$ where every released detector contradicts the GT; they are small relative to the annotated tissue, and excluding them is one of the sensitivity variants of the main text, under which the ranking against the comparators tuned at their released threshold does not change. A point-wise \emph{adjudication} of the TwI references by one reader (not a pathologist) estimated that \AdjGtGlass\,\% of the annotated tissue area is glass.
\input{generated/tab_gt}

\section{Runtime}
Table~\ref{tab:runtime} gives the wall time per slide from the WSI on six test slides, with every method on the same hardware. On all \NumTestSlides{} test slides \Method needs a median of \RtTraceGpuAllMedian\,s on an RTX~3090 (90th percentile \RtTraceGpuAllPninety\,s), of which \RtTraceFmShare\,\% is spent in the foundation model; the time therefore grows with the number of windows sent to it.
\input{generated/tab_runtime}

%% file: generated/tab_groups.tex
\begin{table}[H]
\centering
\footnotesize
\setlength{\tabcolsep}{3.5pt}
\caption{Case groups/slides per source and split. Sources without a patient identifier are grouped per slide, so slides of one patient could fall into different splits.}
\label{tab:groups}
\begin{adjustbox}{max width=\columnwidth}
\begin{tabular}{llrrrr}
\toprule
Source & Case group & DEV & CAL & TEST & External \\
\midrule
ACROBAT & case & 40/79 & 30/60 & 30/60 & -- \\
B\'andi et al. & specimen & -- & -- & -- & 7/10 \\
DLBCL-Morphology & TMA block & 3/18 & 2/12 & 2/12 & -- \\
HISTAI & slide (no patient id) & 1/1 & 1/1 & 1/1 & -- \\
Human Protein Atlas & slide (no patient id) & -- & 1/1 & -- & -- \\
MONKEY & patient & 22/22 & 17/17 & 16/16 & -- \\
MOSTWIEDZY & slide (no patient id) & 13/13 & 10/10 & 9/9 & -- \\
MSKCC & slide (no patient id) & 13/13 & 10/10 & 9/9 & -- \\
NADT & patient & 1/3 & 1/18 & -- & -- \\
TCGA & patient (barcode) & 64/64 & 53/53 & 43/43 & -- \\
TG-GATEs & slide (no patient id) & 4/4 & 3/3 & 3/3 & -- \\
\bottomrule
\end{tabular}
\end{adjustbox}
\end{table}

%% file: generated/tab_source.tex
\begin{table}[H]
\centering
\footnotesize
\setlength{\tabcolsep}{3.5pt}
\caption{Failed/total case groups on the TwI test set by source. Released detectors: threshold and post-processing DEV-tuned; U-Net: seed 0.}
\label{tab:source}
\begin{adjustbox}{max width=\columnwidth}
\begin{tabular}{lrrrrrr}
\toprule
Source & SCOUT & U-Net & AtlasPatch & HEST & nnU-Net & GrandQC \\
\midrule
ACROBAT & 10/30 & 11/30 & 24/30 & 8/30 & 26/30 & 30/30 \\
DLBCL-Morphology & 1/2 & 1/2 & 2/2 & 2/2 & 2/2 & 2/2 \\
HISTAI & 0/1 & 0/1 & 1/1 & 0/1 & 1/1 & 1/1 \\
MONKEY & 0/16 & 0/16 & 3/16 & 5/16 & 4/16 & 1/16 \\
MOSTWIEDZY & 2/9 & 1/9 & 2/9 & 1/9 & 4/9 & 9/9 \\
MSKCC & 2/9 & 2/9 & 4/9 & 6/9 & 5/9 & 9/9 \\
TCGA & 6/43 & 7/43 & 6/43 & 11/43 & 11/43 & 7/43 \\
TG-GATEs & 0/3 & 0/3 & 0/3 & 0/3 & 0/3 & 0/3 \\
\bottomrule
\end{tabular}
\end{adjustbox}
\end{table}

%% file: generated/tab_flag.tex
\begin{table}[H]
\centering
\footnotesize
\setlength{\tabcolsep}{3.5pt}
\caption{Review flag of \Method. $\tau$ is calibrated on CAL with Learn-then-Test ($\delta=\CertDelta$). Silent: a group with a failed slide that is not flagged; the bound controls their rate, not the failure rate among unflagged groups.}
\label{tab:flag}
\begin{adjustbox}{max width=\columnwidth}
\begin{tabular}{rrlrrrrr}
\toprule
$\alpha$, \% & $\tau$ & Set & Slides flagged, \% & Groups flagged, \% & Failed groups flagged, \% & Silent groups & Failure among unflagged groups, \% \\
\midrule
10 & 0.207 & CAL & 25.9 & 28.9 & 50.0 & 6/128 & 6.6 \\
10 & 0.207 & TEST & 31.4 & 31.9 & 61.9 & 11/113 & 10.4 \\
5 & 0.065 & CAL & 64.9 & 64.8 & 83.3 & 2/128 & 4.4 \\
5 & 0.065 & TEST & 71.2 & 69.0 & 95.2 & 3/113 & 2.9 \\
\bottomrule
\end{tabular}
\end{adjustbox}
\end{table}

%% file: generated/tab_gt.tex
\begin{table}[H]
\centering
\footnotesize
\setlength{\tabcolsep}{3.5pt}
\caption{Quality of the reference masks. Suspect region: a component of at least 0.1\,mm$^2$ where all released detectors contradict the GT. Adjudication: stratified point sampling by one reader, area-weighted estimate with approximate 95\% CI.}
\label{tab:gt}
\begin{adjustbox}{max width=\columnwidth}
\begin{tabular}{lrrrr}
\toprule
Dataset & Slides & Suspect missed tissue, \% of GT & Suspect false tissue, \% of GT & Slides with a suspect region, \% \\
\midrule
Tissues with Issues & 555 & 1.21 & 0.27 & 41 \\
B\'andi et al. & 10 & 0.03 & 0.34 & 50 \\
\midrule
\multicolumn{5}{l}{Adjudicated TwI GT error (one reader, 200 points): GT tissue that is glass 8.2\% [0.8, 15.7]; near-tissue GT background that is tissue 0.4\%} \\
\bottomrule
\end{tabular}
\end{adjustbox}
\end{table}

%% file: generated/tab_runtime.tex
\begin{table}[H]
\centering
\footnotesize
\setlength{\tabcolsep}{3.5pt}
\caption{Wall time per slide from the WSI on six TwI test slides, models loaded beforehand (HistoQC includes its start-up). DT: DEV-tuned. CPU: 16 threads of an AMD Ryzen Threadripper PRO 7995WX.}
\label{tab:runtime}
\begin{adjustbox}{max width=\columnwidth}
\begin{tabular}{llrr}
\toprule
Method & Hardware & Median, s & Range, s \\
\midrule
\textbf{SCOUT (ours)} & RTX 3090 & 33.3 & 4.8--77.1 \\
\textbf{SCOUT (ours)} & CPU & 576.6 & 78.9--1462.0 \\
Tuned threshold & CPU & 3.3 & 0.2--3.8 \\
GrandQC (DT) & RTX 3090 & 1.0 & 0.1--2.0 \\
GrandQC (DT) & CPU & 6.1 & 0.4--12.3 \\
HEST (DT) & RTX 3090 & 65.0 & 6.4--105.5 \\
HEST (DT) & CPU & 1302.0 & 178.9--3161.1 \\
AtlasPatch & CPU & 1.3 & 1.1--1.8 \\
HistoQC & CPU & 35.4 & 7.3--108.8 \\
\bottomrule
\end{tabular}
\end{adjustbox}
\end{table}